\documentclass[10pt,twocolumn,letterpaper]{article}

\usepackage[pagenumbers]{cvpr} 

\usepackage{graphicx}
\usepackage{booktabs}
\usepackage{multirow}
\usepackage{makecell}
\usepackage{xcolor}
\usepackage{forloop}
\usepackage{bm}
\usepackage{forloop}
\usepackage{afterpage}
\definecolor{colorfirst}{rgb}{.866,.945, 0.831} 
\definecolor{colorsecond}{rgb}{1, 0.98, 0.83} 
\definecolor{colorthird}{rgb}{0.76, 0.87, 0.92} 

\newcommand{\tocite}[1]{\textcolor{red}{[TOCITE]}}
\newcommand{\toexp}[1]{\textcolor{blue}{[TOEXP]}}
\newcommand{\PAR}[1]{\vskip4pt \noindent{\bf #1~}}

\newcommand{\methodname}{StreetCrafter\xspace}
\newcommand{\methodnameblank}{\methodname\ }

\usepackage[pagebackref,breaklinks,colorlinks]{hyperref}

\def\paperID{3630} 
\def\confName{CVPR}
\def\confYear{2025}

\begin{document}

\title{\methodname: Street View Synthesis with Controllable Video Diffusion Models}

\author{
    Yunzhi Yan$^{1,2^*}$
    \quad Zhen Xu$^{1^*}$
    \quad Haotong Lin$^{1}$
    \quad Haian Jin$^{3}$ 
    \quad Haoyu Guo$^{1}$ \\
    \quad Yida Wang$^{2}$
    \quad Kun Zhan$^{2}$
    \quad Xianpeng Lang$^{2}$
    \quad Hujun Bao$^{1}$ 
    \quad Xiaowei Zhou$^{1}$ 
    \quad Sida Peng$^{1^\dag}$  \\
    $^1$ Zhejiang University \quad
    $^2$ Li Auto Inc. \quad
    $^3$ Cornell University \quad
}

\twocolumn[\maketitle\vspace{0em}\begin{center}
    \vspace{-8mm}
    \captionsetup{type=figure}
    \includegraphics[width=1.0\textwidth]{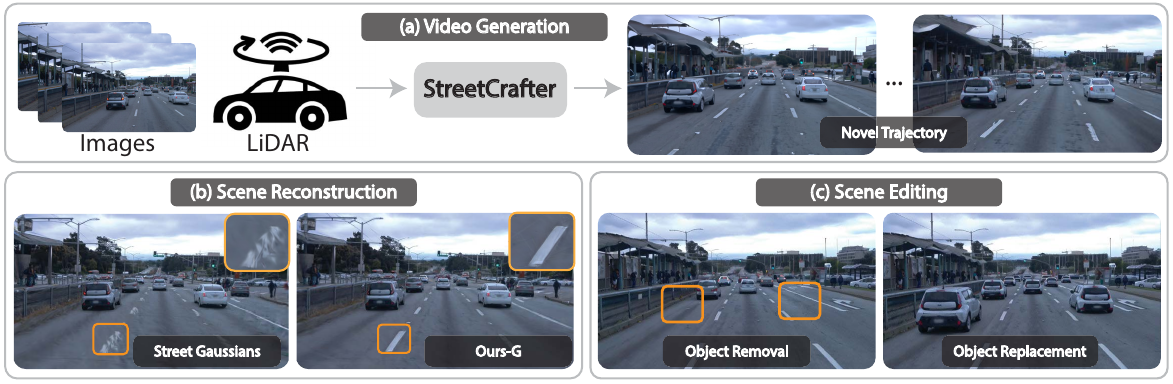}
    \captionof{figure}{%
        \methodnameblank is a novel controllable video diffusion model, which enables precise pose controllability for novel view synthesis in street scenes, using calibrated images and LiDAR as input. 
        \methodnameblank can also serve as data prior to improve the scene reconstruction quality and support scene editing operations without per-scene optimization, such as object removal and replacement.  
    }
    \vspace{-1mm}
    \label{fig:teaser}
\end{center}\bigbreak]
\let\thefootnote\relax\footnotetext{$^*$Equal contribution. $^\dag$Corresponding author.}

\begin{abstract}
This paper aims to tackle the problem of photorealistic view synthesis from vehicle sensor data. 
Recent advancements in neural scene representation have achieved notable success in rendering high-quality autonomous driving scenes, but the performance significantly degrades as the viewpoint deviates from the training trajectory. 
To mitigate this problem, we introduce StreetCrafter, a novel controllable video diffusion model that utilizes LiDAR point cloud renderings as pixel-level conditions, which fully exploits the generative prior for novel view synthesis, while preserving precise camera control. 
Moreover, the utilization of pixel-level LiDAR conditions allows us to make accurate pixel-level edits to target scenes. 
In addition, the generative prior of StreetCrafter can be effectively incorporated into dynamic scene representations to achieve real-time rendering. 
Experiments on Waymo Open Dataset and PandaSet demonstrate that our model enables flexible control over viewpoint changes, enlarging the view synthesis regions for satisfying rendering, which outperforms existing methods.
The code is available at 
\href{https://zju3dv.github.io/street_crafter}{https://zju3dv.github.io/street\_crafter}.
  
\end{abstract}

\newcounter{loopcounter}

\section{Introduction}
\label{sec:intro}

Modeling dynamic street scenes is a crucial step in developing autonomous driving simulators.
Models capable of generating high-quality views can enable closed-loop evaluations of autonomous systems and create corner-case data at a low cost.
The primary challenge is to achieve real-time and high-quality view synthesis across diverse trajectories using vehicle RGB and LiDAR data of a single trajectory.

Recent methods~\cite{kerbl3Dgaussians, mildenhall2020nerf} have achieved great success in novel view synthesis for static scene reconstruction, providing valuable insights for dynamic street modeling. 
Based on this, recent works~\cite{yan2024street, zhou2024drivinggaussian} extend 3DGS~\cite{kerbl3Dgaussians} to dynamic street scenes by modeling moving vehicles through a scene graph.
While these methods enable high-quality, real-time view synthesis, significant artifacts appear in viewpoints that are far from the training trajectory, as shown in Figure~\ref{fig:teaser}.
This issue arises from the insufficient observation in the training input for these regions and the limited ability of view extrapolation for reconstruction-based methods.

Meanwhile, video diffusion models~\cite{gao2024vista, blattmann2023stablevideodiffusionscaling} have demonstrated the ability to generate photorealistic views for novel camera trajectories from just a few input images, leveraging training on large-scale video datasets.
However, these models typically rely on text prompts as control signals, which are high-level instructions and lack fine-grained controllability, thereby hindering their application in autonomous driving simulation.

In this paper, we propose \textit{\methodname}, a novel controllable video diffusion model, which enables precise control over novel view synthesis in street scenes.
Our key observation is that point cloud rendering from LiDAR sensors provides precise geometric information, which despite being incomplete and noisy, can serve as a precise camera pose representation.
To utilize this representation, we incorporate the point cloud rendering as a condition for video diffusion models.
Specifically, we aggregate the colorized LiDAR from adjacent frames to form a global point cloud in world space, which is then rendered to RGB images based on the given camera pose input, serving as a pixel-level pose condition in image space. 
Thanks to the pixel-level condition, we empirically find that even training only on the sequences of single-lane driving data enables high-quality view synthesis across multiple lanes during testing by changing the conditions based on novel view input.
Moreover, the proposed pixel-level condition can be utilized to enable scene editing operations without per-scene optimization, simply by manipulating the LiDAR points as shown in Figure~\ref{fig:teaser}.

However, \methodname faces the challenges of high rendering latency, particularly 0.2fps for a $576 \times 1024$ frame.
This motivates us to further distill \methodname into a dynamic 3DGS~\cite{kerbl3Dgaussians, yan2024street} representation, 
enabling it to perform real-time high-quality view synthesis under large viewpoint changes.
Specifically, we apply \methodname to generate a series of images along the novel trajectory. 
These generated images can serve as extra supervision for the scene representation beyond the original training trajectory inputs.
This distillation process further combines the advantages of 3D scene representation and video diffusion model, achieving state-of-the-art performance in rendering results.

We evaluate our method on Waymo Open Dataset~\cite{Sun_2020_CVPR} and PandaSet~\cite{xiao2021pandaset}. 
The experimental results show that our method outperforms the state-of-the-art methods in terms of image quality, particularly for view extrapolation, while maintaining the ability of real-time rendering.
Our method also enables various scene editing operations without the need for per-scene optimization, such as object removal, replacement and translation. 

Overall, this work makes the following contributions:
\begin{itemize}
  \item We propose a novel controllable video diffusion model, \methodname, which provides precise camera control for novel view synthesis of street scenes. 
  \item We demonstrate that \methodname can be efficiently distilled into a dynamic 3D scene representation, achieving state-of-the-art performance in street view synthesis.
\end{itemize}

\begin{figure*}[ht]
    \centering
    \includegraphics[width=\linewidth]{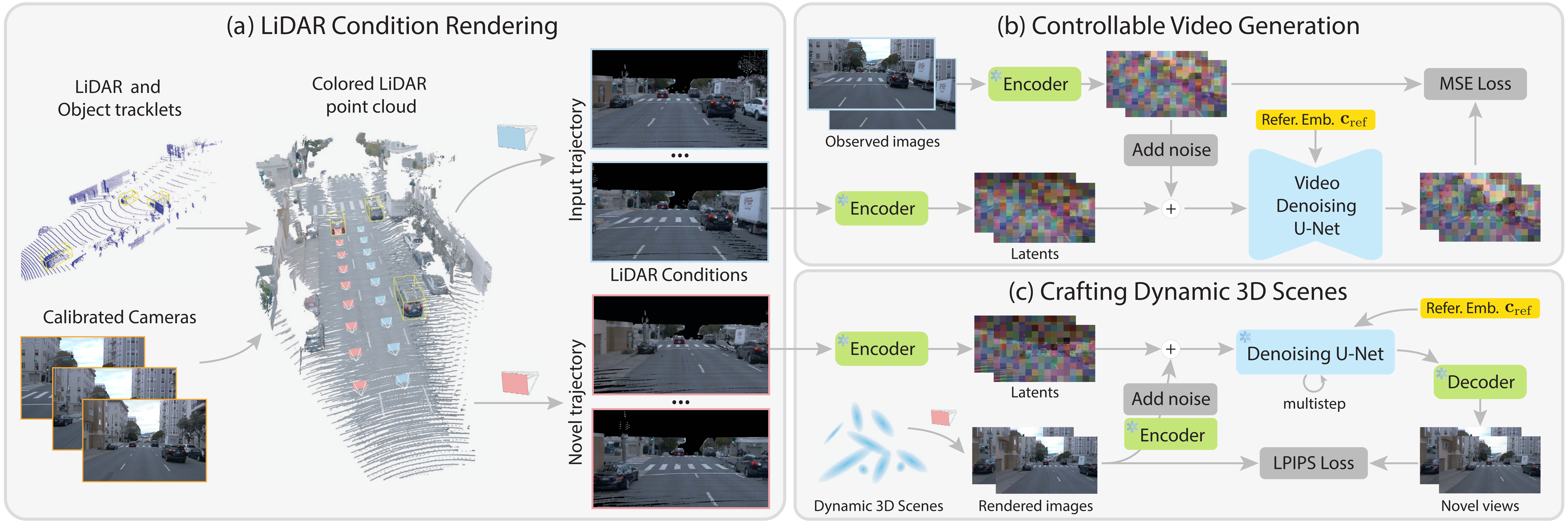}
    \caption{
        \textbf{Overview of \methodname.} 
        (a) We process the LiDAR using calibrated images and object tracklets to obtain a colorized point cloud, which can be rendered to image space as pixel-level conditions. 
        (b) Given observed images and reference image embedding $\mathbf{c}_\text{ref}$, we optimize the video diffusion model conditioned on the LiDAR renderings to perform controllable video generation. 
        (c) Starting from the rendered images and LiDAR conditions under novel trajectory, we use the pretrained controllable video diffusion model to guide the optimization of the dynamic 3DGS representation by generating novel views as extra supervision signals. 
      } 
    \label{fig:pipeline}
    \vspace{-3mm}
\end{figure*}

\section{Related Work}
\PAR{Video Diffusion Models}
VDM~\cite{ho2022video} is the first video diffusion model that applies a diffusion model with a space-time factorized U-Net to a video generation task.
Imagen-Video~\cite{ho2022imagen} proposes cascaded diffusion models to achieve higher resolution.
\cite{blattmann2023align,he2022latent,wang2023modelscope,wang2023lavie,zhou2022magicvideo} demonstrate learning VDM in the latent space, enabling high-resolution video generation at low computational cost.
\cite{wang2023drivedreamer,gao2023magicdrive,gao2024vista} learn specific elements and policies from street-view video data, enabling it to generate realistic street-view videos.

To better support downstream applications such as reconstruction, some works have proposed camera-controlled video generation methods.
\cite{you2024nvs,hou2024training,xiao2024video} propose training-free methods that control the denoising process of an existing diffusion model to achieve controllable video generation.
While these methods do not require training or fine-tuning the diffusion model, their performance is limited due to ambiguity during generation.
Some recent works finetune the video diffusion model with additional camera parameters as additional input.
\cite{voleti2024sv3d,kwak2024vivid,melaskyriazi2024im3d} generate novel views for object-centric scene.
Some methods achieve camera control by injecting camera parameters into the video diffusion model \cite{xu2024camco,wang2024motionctrl,watson2024controllingspacetimediffusion,bahmani2024vd3d,vanhoorick2024gcd} based on the architecture of U-Net or transformer~\cite{peebles2023scalable}.
Another line of works use geometry foundation models \cite{wang2024dust3r,lin2024promptda} to build explicit representation such as point cloud as guidance to the video diffusion model~\cite{liu2024reconx,yu2024viewcrafter,muller2024multidiff}. 
However, they mainly handle static scenes, while our method leverages the more accurate LiDAR prior to control complex driving scenes.

\PAR{Street Scene Representation}
NeRF~\cite{mildenhall2020nerf} and 3DGS~\cite{kerbl3Dgaussians} have become the leading approaches for modeling autonomous driving scenes. Block-NeRF~\cite{tancik2022block} uses a block-based modeling approach to represent large-scale street scenes. 
Considering that street scenes typically include moving elements such as vehicles and pedestrians, 
\cite{turki2023suds,yang2023emernerf, chen2023periodic} encode time as an additional input to build 4D representation.
\cite{ost2021neural, wu2023mars, fischer2024dynamic, yang2023unisim, tonderski2024neurad, yan2024street, zhou2024drivinggaussian, ml-nsg, chen2024omnire} decompose the scene into moving objects and static backgrounds, which are reconstructed separately and combined in world space by tracking the moving objects at each time step.

There have been works that try to utilize the LiDAR input~\cite{chang2023neural,ost2022neural,sun2025pointnerf++,lu2023urban,sun2024lidarf}
to enhance the model's capability of capturing scene geometry and generalizing to novel viewpoints.
However, they mainly deal with static scenes only using supervision from the input trajectory, 
while our model combines the LiDAR with generative prior to generate guidance on novel trajectories for dynamic urban scenes.

\PAR{Reconstruction with Diffusion Prior}
\cite{lin2023magic3d, poole2022dreamfusion} use SDS for lifting 2D generation to 3D, achieving text-based 3D generation.
\cite{liu2023one2345, li2023instant3d, long2024wonder3d} generate multi-view predictions based on a single-view image input using diffusion model, and then use multi-view reconstruction methods to obtain the reconstructed 3D model.
\cite{wu2024reconfusion, gao2024cat3d} further extend the diffusion prior to scene level.
Some works have utilized video diffusion models to generate novel views with improved multi-view consistency compared to image diffusion models.
\cite{liu2024reconx, yu2024viewcrafter} employ DUSt3R~\cite{wang2024dust3r} to build coarse geometry structure from sparse viewpoints, which serves as guidance for the video diffusion model, allowing it to produce additional inputs for reconstructing the static scene.
\cite{chen2024mvsplat360,liu20243dgsenhancer} use the degraded rendering results as condition to perform sparse view synthesis with video diffusion models.
Recent works \cite{yu2024sgd,han2024ggs,gao2024magicdrive3d, zhao2024drive, hwang2024vegs} enhance street scene reconstruction quality by incorporating diffusion prior to generate additional views, similar to our methods. However, their diffusion models lack precise camera control, which limits the accuracy and visual quality of their reconstruction.

\section{Method} \label{method}
Given a recorded autonomous driving scene with calibrated images, LiDAR point clouds, and corresponding object tracklets, 
our goal is to develop a model capable of synthesizing photorealistic images for novel views. 
We first give a brief introduction to video diffusion models and 3D Gaussian Splatting in Section \ref{sec:preliminaries}.
Then, we introduce our controllable video diffusion model \methodnameblank in Section \ref{sec:street_crafter_video}. Finally, Section \ref{sec:street_crafter_gaussian}
discusses how to distill \methodnameblank into a 3D representation for real-time rendering.

\subsection{Preliminaries}
\label{sec:preliminaries}
\paragraph{Video Diffusion Models.}

Diffusion models~\cite{song2021scorebased, ho2020denoising} have become a frontline methodology for video generation 
in recent years. These models learn the underlying data distribution by a forward process and a reverse process.
In the forward diffusion process, Gaussian noise $\epsilon \sim \mathcal{N}(0, 1)$ is incrementally added to the initial latent $x_0 \sim p(x)$, resulting
in the noisy latent $x_t$:
\begin{equation}
    \label{eq:add_noise}
    x_t = \sqrt{\bar{\alpha}_t} x_0 + \sqrt{1 - \bar{\alpha}_t} \epsilon,
\end{equation}
where $t \in \{1, \dots, T\}$ is the diffusion timestep and $\bar{\alpha}_t \in (0, 1]$ is the noise scheduling parameter.
In the reverse diffusion process, the model learns to iteratively denoise the latent to clean data with a trained network 
$\mathcal{F}_\theta(x_{t-1} | x_t)$. We build our model based on Vista~\cite{gao2024vista}, 
which is a driving world model finetuned from Stable Video Diffusion (SVD)~\cite{blattmann2023stablevideodiffusionscaling} following the
continuous-timestep formula~\cite{Karras2022edm}. Given conditional input image $c$, the network $\mathcal{F}_\theta$ is optimized by the loss function:
\begin{equation}
    \mathcal{L} = \mathbb{E}_{x_0, \epsilon, c, t} \left[ \| x_0 - \mathcal{F}_\theta (x_t, t, c) \|_2^2 \right].
\end{equation}

\paragraph{3D Gaussian Splatting.} 
3DGS ~\cite{kerbl3Dgaussians} represents the scene using a set of anisotropic Gaussian defined in the 3D world. Each Gaussian $\mathcal{G}$ is assigned with 
opacity $o \in \mathbb{R}$, spherical harmonics (SH) coefficient $\bm{z} \in \mathbb{R}^{k}$, position vector $\bm{\mu} \in \mathbb{R}^{3}$, rotation quaternion $\bm{q} \in \mathbb{R}^{4}$ and scale factor $\bm{s} \in \mathbb{R}^{3}$. 
The Gaussian kernel distribution is formulated as:
\begin{equation}
    \label{eq:gaussian}
    \begin{aligned}
        \mathcal{G}(\bm{x}) = \exp({-\frac{1}{2} (\bm{x}-\bm{\mu})^\top \bm{\Sigma}^{-1}(\bm{x}-\bm{\mu})}),
    \end{aligned}
\end{equation}
where $\bm{\Sigma}=\bm{R}\bm{S}\bm{S}^T\bm{R}^T$, $\bm{S}$ is the scaling matrix determined by $\bm{s}$ and $\bm{R}$ is the rotation matrix determined by $\bm{q}$.
Given camera extrinsic $\mathbf{W}$ and intrinsic $\mathbf{K}$, 2D covariance matrix $\bm{\Sigma}^{*}$ in screen space is computed as~\cite{zwicker2001ewa}:
\begin{equation}
    \label{eq:ewa_splatting}
    \bm{\Sigma}^{*} = \mathbf{J} \mathbf{W} \bm{\Sigma} \mathbf{W}^T  \mathbf{J}^T.
\end{equation}
The color $\bm{C}$ of each pixel is rendered by alpha compositing the view-dependent color $\bm{c}$ in depth order:  
\begin{equation}
    \label{eq:rendering}
    \bm{C} = \sum_{i \in N} \bm{c}_i \alpha_i \prod_{j=1}^{i-1} (1 - \alpha_j).
\end{equation}

\subsection{Controllable Video Generation}
\label{sec:street_crafter_video}

In this section, we seek to build a diffusion model that takes a reference image $\mathbf{I}_{\text{ref}}$ and a set of camera trajectories $\{\mathbf{C}_{i}\}_{i=1}^{K}$ to generate the same number of video frames. 
Previous methods~\cite{wang2024motionctrl,he2024cameractrl} typically rely on camera poses as control signals, which are insufficient for street scenes with complex backgrounds and multiple moving objects.
To tackle this problem, we propose a novel controllable video diffusion model \methodnameblank, which leverages the LiDAR input to provide precise control over viewpoint change during the diffusion denoising process.

\paragraph{Building LiDAR condition.}
Considering a driving scene with $N$ recorded frames, we first project the LiDAR point cloud onto the calibrated image plane and colorize it by querying the pixel value. 
Then we utilize the object tracklets to separate the object point cloud from the background, resulting in the background point cloud $\{\mathbf{P}_{i}^{b}\}_{i=1}^{N}$
and object point cloud $\{\mathbf{P}_{i}^{o}\}_{i=1}^{N}$, which are defined in the canonical bounding box coordinate system for each dynamic instance $o$. 
Given the input camera pose $\mathbf{C}_i$ at frame $t_i$, we aggregate the LiDAR points within a temporal window of size $l$ to form a unified point cloud $\mathbf{P}$ in the world coordinate system.
The object point cloud is warped into the world coordinate system using the object pose $\mathbf{T}_{o}^{t_i}$.

Since there still exist numerous missing and occluded regions when treating each point cloud as a pixel in the camera screen, we assign a fixed radius to each LiDAR point in NDC space
and perform point rasterization under the camera pose $\mathbf{C}_i$ instead of directly projecting $\mathbf{P}$ onto the image plane, yielding the LiDAR condition image $\mathbf{I}^{c}_{i}$ as shown in Figure~\ref{fig:pipeline}.
In this way, we establish a connection between the novel camera trajectories and input calibrated images in a pixel-wise manner by utilizing LiDAR as coarse scene geometry. 
In comparison with conditional signals, such as camera pose embedding, the LiDAR condition image could provide much stronger guidance as the network only needs to recover clean images from noisy input conditions rather than learning the complicated process of converting camera parameters into video frames.

\paragraph{Training and inference.}
To train the model with input images $\{\mathbf{I}_{i}\}_{i=1}^{K}$ and camera trajectory $\{\mathbf{C}_{i}\}_{i=1}^{K}$, we choose the first frame $\mathbf{I}_0$ as reference image $\mathbf{I}_{\text{ref}}$ and obtain the corresponding LiDAR conditions $\{\mathbf{I}^{c}_{i}\}_{i=1}^{K}$, as shown in Figure~\ref{fig:pipeline}. 
The inputs and LiDAR conditions are further encoded into latent space as $\{\mathbf{z}_{i}\}_{i=1}^{K}$ and $\{\mathbf{z}^{c}_{i}\}_{i=1}^{K}$ with a pre-trained VAE encoder.
We then inject $\mathbf{z}^{c}_{i}$ into the first layer of the U-Net architecture by adding a trainable zero convolutional layer~\cite{zhang2023adding}  $\Theta_z$ to $\mathcal{F}_\theta$ and perform element-wise addition:
\begin{equation}
    \label{eq:condition}
    \hat{\mathbf{z}}_{i, t} = \mathbf{z}_{i, t} + \mathcal{Z}(\mathbf{z}^{c}_{i} \, ; \, \Theta_z),
\end{equation}
where $\mathcal{Z}(\cdot \, ; \, \Theta_z)$ indicates the zero convolutional layer and $\mathbf{z}_{i, t}$ is the noisy latent from $\mathbf{z}_{i}$ and diffusion timestep $t$ using Eq. \ref{eq:add_noise}. 
We find that this minor modification could provide sufficient guidance without introducing additional computation costs.
The video denoising U-Net $\mathcal{F}_\theta$ is optimized by minimizing the following loss objective:
\begin{equation}
    \label{eq:condition_diffusiion_loss}
    \mathcal{L} = \mathbb{E}_{\mathbf{z}_{i}, \epsilon, \mathbf{c}_{\text{ref}}, , c_{\text{p}, t}} \left[\| \mathbf{z}_{i} - \mathcal{F}_\theta (\hat{\mathbf{z}}_{i, t}, t, \mathbf{c}_{\text{ref}}, \mathbf{c}_{\text{p}}) \|_2^2 \right],
\end{equation}
where $\mathbf{c}_\text{ref}$ and $\mathbf{c}_\text{p}$ refer to the reference image CLIP embedding~\cite{radford2021learning} and LiDAR conditions respectively. 

During inference with novel camera trajectory $\{\hat{\mathbf{C}}_{i}\}_{i=1}^{K}$ as input, we select the input camera closest to $\hat{\mathbf{C}}_{1}$ as reference image $\mathbf{I}_\text{ref}$ and render the novel view LiDAR conditions denoted as $\{\hat{\mathbf{I}}^{c}_{i}\}_{i=1}^{K}$.  
Starting from sampled noise $\epsilon \sim \mathcal{N}(0, I)$, we iteratively denoise the noisy latent $\hat{\mathbf{z}}_{i}$ with the trained denoising network $\mathcal{F}_\theta$ conditioned on $\{\hat{\mathbf{I}}^{c}_{i}\}_{i=1}^{K}$ and $\mathbf{I}_\text{ref}$ into clean latents, which is further
decoded into novel view images $\{\hat{\mathbf{I}}_{i}\}_{i=1}^{K}$ with a pre-trained VAE decoder.

\subsection{Crafting Dynamic 3D Scenes}
\label{sec:street_crafter_gaussian}

In this section, our goal is to distill the generative prior of the controllable video diffusion model into a more consistent 3DGS~\cite{kerbl3Dgaussians} representation for real-time rendering.
The main drawback of 3DGS is that it cannot generalize well to novel viewpoints away from input cameras, which is a common problem for reconstruction-based methods.
To address this issue, we propose a novel framework that leverages our video diffusion model to apply constraints to 3DGS along novel trajectories during optimization.

\paragraph{Scene representation. }
To model dynamic urban scenes with 3DGS, we follow existing approaches~\cite{yan2024street,chen2024omnire} and use a distinct set of Gaussian parameters to model the background and each foreground moving object. 
The object Gaussians $\mathcal{G}_v$ is defined in the canonical coordinate system determined by the object tracklets. Given the SE(3) pose $\bm{T}_{v}^{t} = (\bm{R}_{v}^{t},  \bm{t}_{v}^{t})$ at timestamp $t$, $\mathcal{G}_v$
can be mapped into the world coordinate system for global rendering as:
\begin{equation}
    \label{eq:streetgs}
    \hat{\bm{\mu}}_{v} = \bm{R}_{v}^{t} \bm{\mu}_{v} + \bm{t}_{v}^{t}, \quad \hat{\bm{R}}_{v} = \bm{R}_{v}^{t} \bm{R}_{v},
\end{equation}
where $\bm{\mu}_{v}, \bm{R}_{v}$ and $\hat{\bm{\mu}}_{v}, \hat{\bm{R}}_{v}$ denotes the position and rotation of $\mathcal{G}_v$ in local and world coordinate system, respectively.
The distant region of the scene is modeled by a high-resolution cubemap, which is combined with the rendered color $\bm{C}_{\mathcal{G}}$ from Eq.~\ref{eq:rendering} using rendered opacity $\bm{O}_{\mathcal{G}}$.

\paragraph{Novel view generation.}
To better align the generated video frames from the diffusion model with the 3DGS scene representation, 
we generate samples from the noisy rendered latent instead of pure noise inspired by previous works~\cite{wu2024reconfusion,poole2022dreamfusion}. 
We find this can help maintain the overall scene structure and accelerate the training process due to fewer denoising steps.    
To be specific, we render images $\{{\mathbf{I}}_i^r\}_{i=1}^{K}$ from novel views $\{\hat{\mathbf{C}}_{i}\}_{i=1}^{K}$, which are encoded and perturbed to noisy latents $\{\mathbf{z}_{i, t}^r\}_{i=1}^{K}$ given diffusion timestep $t$ derived from noise scale $s$.
Then we iteratively denoise the latents to clean samples by running the EDM sampling~\cite{Karras2022edm} used in Vista. The samples are further decoded to novel view images $\{\hat{\mathbf{I}}_{i}\}_{i=1}^{K}$ for supervision.
Based on the following generation process, we employ a progressive optimization strategy by gradually reducing the noise scale $s$. 
This helps the model learn to remove artifacts by relying more on the diffusion prior in the early training stage and progressively focusing on refining details as the training advances.

\paragraph{Loss function.}
We construct our training set by combining input views with novel views generated by the video diffusion model. 
In each training iteration, we randomly sample a camera $\mathbf{C}$ with the ratio of selecting a novel view camera set to $p$.
The Gaussian scene representation $\mathcal{G}$ is optimized using the following loss function:
\begin{equation}
    \label{eq:gaussian_loss}
    \begin{aligned}
        \mathcal{L}_{\text{input}} = \lambda_{1} \mathcal{L}_1 + & \lambda_{\text{ssim}} \mathcal{L}_{\text{ssim}} + \lambda_{\text{lpips}} \mathcal{L}_{\text{lpips}} + \mathcal{L}_{\text{g}}, 
        \\ 
        \mathcal{L}_{\text{novel}} &= \lambda_{\text{novel}} \mathcal{L}_{\text{lpips}},
        \\
    \end{aligned}
\end{equation}
where $\mathcal{L}_1$, \(\mathcal{L}_{\text{ssim}}\), and \(\mathcal{L}_{\text{lpips}}\) denote the \(L_1\), SSIM, and LPIPS losses, respectively.
we select either $\mathcal{L}_{\text{input}}$ or $\mathcal{L}_{\text{novel}}$ as the loss function depending on whether $C$ is a novel view.
In comparison with the original loss function of 3DGS, we additionally introduce the LPIPS~\cite{Zhang_2018_CVPR} loss between the rendered and generated image of novel view as it emphasizes high-level semantic similarity rather than photometric consistency. 
We also add an extra loss $\mathcal{L}_{g} $ for all input views, which includes LiDAR depth loss $\mathcal{L}_{\text{depth}}$, sky mask loss $\mathcal{L}_{\text{sky}}$ and moving objects regularization loss $\mathcal{L}_{\text{reg}}$ to further enhance the scene geometry.
Please refer to the supplementary material for more details on each loss term. 

\section{Implementation Details}\label{sec:implementation}
We initialize \methodnameblank from the pretrained checkpoint of Vista \cite{gao2024vista}. 
We first train all the parameters of the video denoising U-Net at the resolution of 320 $\times$ 576 with batch size of 16 and learning rate of $5e^{-5}$ for 30000 iterations.
Then we fix the temporal layers and finetune the spatial layers of U-Net at the resolution of 576 $\times$ 1024 with batch size of 8 and learning rate of $1e^{-5}$ for another 3000 iterations.
During training, we randomly drop the reference image and LiDAR conditions with a probability of 15\% independently. It takes 2 days to train the model on 8 NVIDIA A800 80GB GPUs using Adam~\cite{kingma2014adam} optimizer. 
During inference, we set the sampling steps to 50 and classifier-free guidance (CFG) \cite{ho2022classifier} scale to 2.5 and generate videos of length $n=25$ at the resolution of $576 \times 1024$.
For novel trajectories longer than $n$, we iteratively sample $n$-length video frames with an overlapping frame length of 5 to construct the full-length video.
The temporal window size $l$ is set to cover the LiDAR point cloud from adjacent $\pm 1s$ frames.

During the distillation process from \methodnameblank to 3DGS, we follow the setting of Street Gaussians~\cite{yan2024street} and train the model for 30000 iterations.
The novel trajectory is built by shifting the input cameras lateral to the heading direction of the ego vehicle for 3 meters across the sequence and we sample the novel view cameras with the ratio $p$ set to 0.4.
The coefficients $\lambda_{1} $, $\lambda_{\text{ssim}}$, $\lambda_{\text{lpips}}$ and $\lambda_{\text{novel}}$ are set to 0.2, 0.8, 0.5 and 0.1, respectively.
The training takes about 1.5 hours on one A800 GPU.

\section{Experiments}

\subsection{Experiments Setup}

We evaluate the performance of \methodnameblank and the crafted dynamic 3DGS representation which we refer to as Ours-V and Ours-G respectively on the task of novel view synthesis.
For Ours-V, we crop and resize the input image to 576 $\times$ 1024 to match the output resolution of video diffusion model during evaluation.
\PAR{Datasets.}
We conduct experiments on Waymo Open Dataset~\cite{Sun_2020_CVPR} and PandaSet~\cite{xiao2021pandaset}, using their 10Hz front camera and synchronized LiDAR.
We select 15 sequences of approximately 100 frames from the validation set of Waymo and 5 sequences of 80 frames from PandaSet to test the novel view synthesis results.
We uniformly sample half of the images in each sequence as the testing frames and use the remaining for training.
Input image resolution is set to $1066 \times 1600$ and $900 \times 1600$ for Waymo and PandaSet, respectively.
The training set of Waymo and the remaining PandaSet sequences are used to train \methodname, resulting in a total of approximately 35,000 training samples.
More details can be found in the supplementary.
 
\PAR{Baselines.}
We compare our method with 3DGS~\cite{kerbl3Dgaussians}, Street Gaussians~\cite{yan2024street}, EmerNeRF~\cite{yang2023emernerf}, UniSim~\cite{yang2023unisim} and NeuRAD~\cite{tonderski2024neurad}.
Street Gaussians~\cite{yan2024street} models the background and each moving object using separate Gaussian models. EmerNeRF~\cite{yang2023emernerf} stratifies scenes into static and dynamic fields, each modeled with a hash grid \cite{mueller2022instant}.
UniSim~\cite{yang2023unisim} and NeuRAD~\cite{tonderski2024neurad} utilize neural feature grids to model dynamic driving scenes with CNN renderer to enhance the ability of view extrapolation.
We enhance 3DGS by incorporating LiDAR depth supervision and LiDAR point cloud initialization. For the rest of the methods, we evaluated the results based on their official implementations.

\begin{figure*}
    \centering
    \includegraphics[width=1\linewidth]{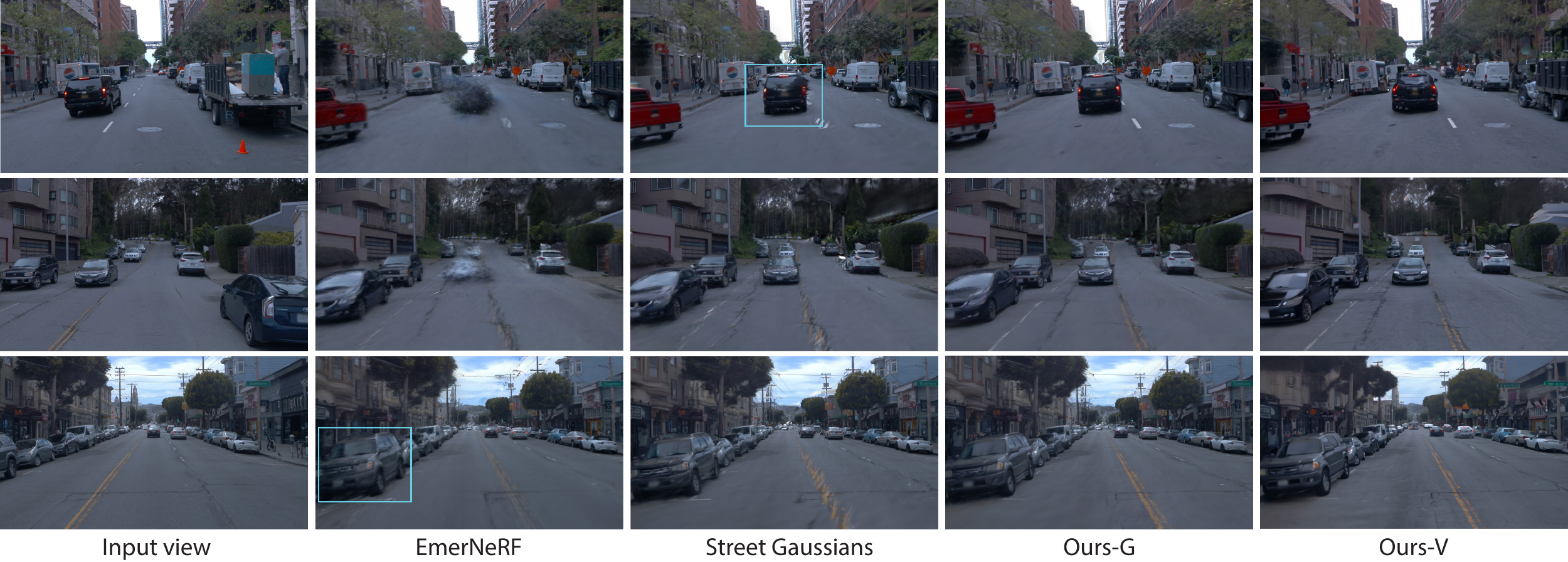}
    \caption{
        \textbf{Qualitative comparisons on the Waymo~\cite{Sun_2020_CVPR} dataset}.
        The camera is laterally shifted for 3 meters to left or right. Input view refers to the closest training camera.
        Ours-G denotes the dynamic 3DGS distilled from \methodname and Ours-V denotes \methodname.  
    }
    \label{fig:waymo_comparison}
\end{figure*}

\begin{figure*}
    \centering
    \includegraphics[width=1\linewidth]{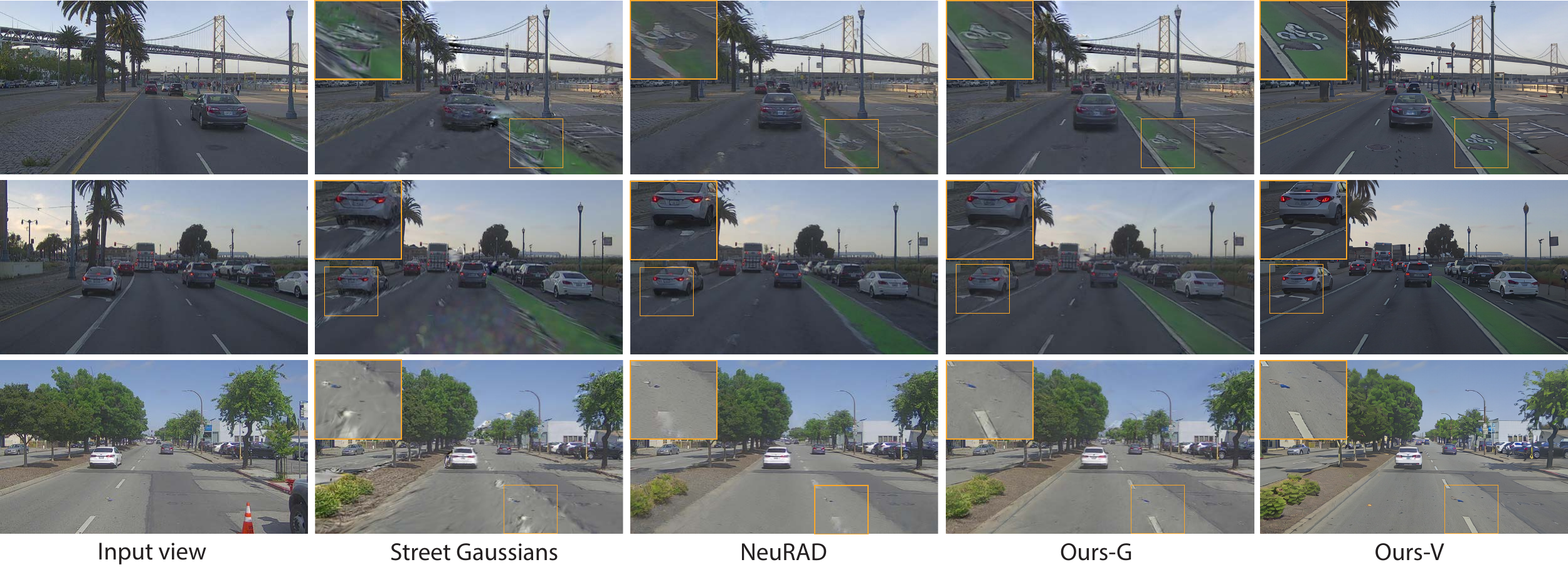}
    \caption{
        \textbf{Qualitative comparisons on the PandaSet~\cite{xiao2021pandaset} dataset}.
        The camera is laterally shifted for 3 meters to left or right.
    }
    \label{fig:pandaset_comparison}
\end{figure*}

\begin{table}[ht]
    \centering
    \scalebox{0.7}{
    \begin{tabular}{cccccc}
        \toprule
        \multirow{2}{*}{Methods} & \multicolumn{2}{c}{Interpolation} & \multicolumn{2}{c}{Lane Shift} & \multirow{2}{*}{FPS} \\
        \cmidrule(lr){2-3} \cmidrule(lr){4-5}
        & PSNR$\uparrow$ & LPIPS$\downarrow$ & FID$\downarrow$ @ 2m & FID$\downarrow$ @ 3m \\
        \midrule
        Ours-V & 27.19 & 0.087 & 62.43 & 73.49 & 0.16 \\
        \midrule
        3DGS~\cite{kerbl3Dgaussians} & 28.85 & 0.148 & 100.50 & 122.52 & \textbf{213.37} \\
        EmerNeRF~\cite{yang2023emernerf} & 26.09 & 0.199 & 89.98 & 110.78 & 0.20 \\
        Street Gaussians \cite{yan2024street} & \textbf{30.95} & 0.130 & 71.42 & 93.38 & 92.16 \\
        Ours-G & 30.05 & \textbf{0.054} & \textbf{58.17} & \textbf{71.40} & 113.16 \\
        \bottomrule
    \end{tabular}
    }
    \caption{
        \textbf{Quantitative results on the Waymo~\cite{Sun_2020_CVPR} dataset.} The rendering image resolution is 1066 $\times$ 1600.
         Ours-G denotes the dynamic 3DGS distilled from \methodname and Ours-V denotes \methodname tested under the resolution of 576 $\times$ 1024.  
    }
    \label{tab:comparison_waymo}
\end{table}


\begin{table}[ht]
    \centering
    \scalebox{0.7}{
    \begin{tabular}{cccccc}
        \toprule
        \multirow{2}{*}{Methods} & \multicolumn{2}{c}{Interpolation} & \multicolumn{2}{c}{Lane Shift} & \multirow{2}{*}{FPS} \\
        \cmidrule(lr){2-3} \cmidrule(lr){4-5}
        & PSNR$\uparrow$ & LPIPS$\downarrow$ & FID$\downarrow$ @ 2m & FID$\downarrow$ @ 3m \\
        \midrule
        Ours-V & 26.10 & 0.090 & 69.68 & 81.99 & 0.19 \\
        \midrule
        3DGS~\cite{kerbl3Dgaussians} & 26.11 & 0.135 & 127.74 & 143.94 & 86.98 \\
        UniSim~\cite{yang2023unisim}  & 25.62 & 0.121 & 75.26 & 92.65 & 6.43 \\
        NeuRAD~\cite{tonderski2024neurad} & 27.00 & 0.098 & 64.65 & 86.44 & 6.01 \\
        Street Gaussians \cite{yan2024street} & \textbf{27.54} & 0.109 & 69.87 & 90.41 & \textbf{88.47} \\
        Ours-G & 26.68  & \textbf{0.062} & \textbf{62.15} & \textbf{78.88} & 80.56 \\
        \bottomrule
    \end{tabular}
    }
    \caption{
        \textbf{Quantitative results on the PandaSet~\cite{xiao2021pandaset} dataset}. The rendering image resolution is 900 $\times$ 1600. }
    \label{tab:comparison_pandaset}
\end{table}


\begin{figure*}
    \centering
    \includegraphics[width=1\linewidth]{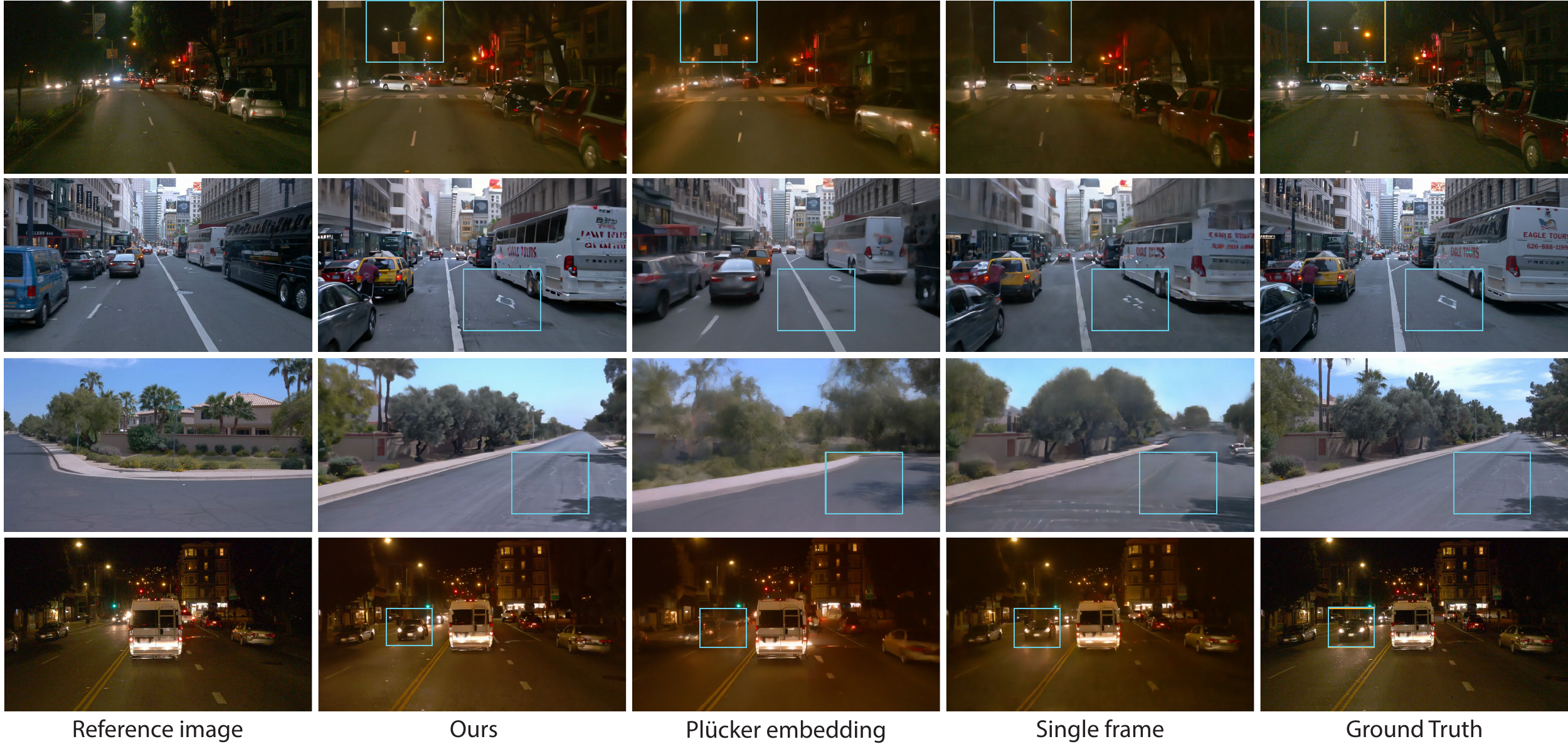}
    \caption{
        \textbf{Visual ablation results on the design choice of \methodname.} 
        The results indicate that our LiDAR condition can provide more accurate control for street view synthesis even when the viewpoint deviates greatly from the reference image.
    }
    \label{fig:ablation_video}
\end{figure*}
\subsection{Comparisons Results}
Tables \ref{tab:comparison_waymo}, \ref{tab:comparison_pandaset}  present the comparison results of our method with baseline methods~\cite{kerbl3Dgaussians,yan2024street,yang2023unisim,tonderski2024neurad,yang2023emernerf} in terms of rendering quality and rendering speed.
We assess the rendering quality under the setting of view interpolation and extrapolation.
We adopt PSNR and LPIPS~\cite{Zhang_2018_CVPR} as evaluation metrics for view interpolation and 
report FID~\cite{heusel2017gans} under the setting of lane shift for view extrapolation since no ground truth image is available.
Our method achieves state-of-the-art performance in extrapolation scenarios while maintaining comparable results for interpolation.
%

Figures \ref{fig:waymo_comparison}, \ref{fig:pandaset_comparison} display the qualitative differences.
Baseline methods tend to generate blurry results with artifacts, particularly in challenging and under-observed regions such as lanes and moving vehicles, while both Ours-G and Ours-V can both generate high-fidelity novel views, thanks to the LiDAR conditions, diffusion prior, and distillation process.
Moreover, Ours-G can achieve rendering speed similar to Street Gaussians~\cite{yan2024street} but demonstrates superior large-view generalization capability as shown in the FID metric.

\subsection{Ablation Studies} 
We first analyze the design choice of \methodnameblank presented in Section \ref{sec:street_crafter_video} with 
several variants on two types of settings from the validation set of Waymo~\cite{Sun_2020_CVPR} dataset as shown in Table~\ref{tab:ablation_condition}.
For the random set, we randomly select 40 video clips from 10 uniformly sampled Waymo sequences. We further specifically choose 10 video clips from which include
complex behaviors such as turnings and lane changes to form the hard set. All the videos have a frame length of 25 with the first frame selected as the reference image
and all the variants are trained under the same setting as our model.
Finally, we ablate on several optimization strategies during the distillation process presented in Section \ref{sec:street_crafter_gaussian}.



\begin{table}[t!]
    \centering
    \scalebox{0.65}{
        \begin{tabular}{lcccccccc}
            \toprule
            \multirow{2}{*}{Methods} & \multicolumn{3}{c}{Random set} & \multicolumn{3}{c}{Hard set} \\
            \cmidrule(lr){2-4} \cmidrule(lr){5-7}
            & PSNR $\uparrow$ & LPIPS $\downarrow$ & FID $\downarrow$ & PSNR $\uparrow$ & LPIPS $\downarrow$ & FID $\downarrow$ \\
            \midrule
            (1) Pl\"ucker embedding & 22.88 &  0.233 & 92.75 & 17.59 & 0.421 & 149.35 \\
            (2) Pose + 3D box & 24.54 & 0.201 & 90.36 & 18.81 & 0.395 & 159.86 \\
            (3) Single frame LiDAR & 25.57 & 0.156 & 73.25 & 20.62 & 0.272 & 106.01 \\
            (4) Projected LiDAR  & 26.15  & 0.135 & 66.29 & 21.57 & 0.246 & 100.51 \\
            (5) Ours & \textbf{27.00} & \textbf{0.121} & \textbf{55.53} & \textbf{22.23} & \textbf{0.218} & \textbf{77.03} \\
            \bottomrule
        \end{tabular}
    }

    \caption{
        \textbf{Ablation study on the design choice of \methodname.} 
        Metrics are averaged over all sampled video clips and our complete model achieves the best performance.  
        }
    \label{tab:ablation_condition}
\end{table}

\begin{figure*}
    \centering
    \includegraphics[width=1\linewidth]{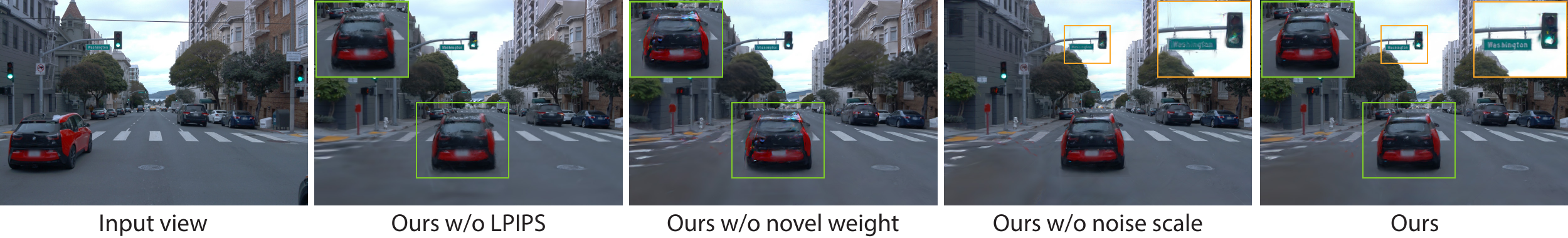}
    \caption{
        \textbf{Visual ablation results on the distillation of \methodname}. The camera is laterally shifted for 3 meters to the left. Our framework refines the texture details
        of nearby regions while maintaining the 3D structure in areas without LiDAR conditions. 
    }
    \vspace{-1em}

    \label{fig:ablation_gaussian}
\end{figure*}

\PAR{Our model with camera pose condition.}
We compare \methodnameblank with two variants conditioned on camera pose as shown in Rows 1-2 of Table~\ref{tab:ablation_condition}.  
We first replace the LiDAR condition with Pl\"ucker embeddings~\cite{sitzmann2021light} of camera rays as the pose representation. 
Although ray embedding serves as a pixel-level condition, it fails to build relationships between the camera pose and the scene geometry.
As shown in Figure~\ref{fig:ablation_video}, the model lacks precise control and results in blurriness as the viewpoint diverges from the reference image.
We then treat the camera parameter as a vector and inject it into the temporal attention layer of the denoising U-Net~\cite{wang2024motionctrl}.
To better model object motions, we replace our LiDAR condition with projected object 3D bounding box similar to recent world models~\cite{wang2023drivedreamer,zhao2024drive}. 
As shown in Table~\ref{tab:ablation_condition}, the experimental results indicate that it lacks controllability due to the complexity of the driving scene. 

\begin{table}[ht]
    \centering
    \scalebox{0.7}{
    \begin{tabular}{lccccc}
        \toprule
        \multirow{2}{*}{Methods} & \multicolumn{2}{c}{Interpolation} & \multicolumn{2}{c}{Lane Shift} \\
        \cmidrule(lr){2-3} \cmidrule(lr){4-5}
        & PSNR$\uparrow$ & LPIPS$\downarrow$ & FID$\downarrow$ @ 2m & FID$\downarrow$ @ 3m \\
        \midrule
        (1) Ours w/o LPIPS loss & 30.80 & 0.108 & 69.16 & 79.63\\ 
        (2) Ours w/o novel weight & 29.38 & 0.054 & 71.97 & 82.11 \\ 
        (3) Ours w/o noise scale & 30.86 & 0.044 & 65.80 & 77.59 \\
        (4) Ours & \textbf{30.88} & \textbf{0.043} & \textbf{64.70} & \textbf{73.45} \\ 
        \bottomrule
    \end{tabular}
    }
    \caption{
        \textbf{Ablation study on the distillation of \methodname.} Metrics are averaged over the selected sequences and our complete model achieves the best performance.  
    }
    \label{tab:ablation_gaussian}
\end{table}

\PAR{The influence of rendering aggregated LiDAR.}
We compare \methodnameblank with two variants conditioned on LiDAR shown in Rows 3-4 of Table~\ref{tab:ablation_condition}.  
We first change the aggregated LiDAR point cloud to single-frame LiDAR input. 
As shown in Figure~\ref{fig:ablation_video}, the generated video lacks fidelity in representing scene texture details, such as the icons on the road since the single-frame LiDAR is extremely sparse even rendered with radius in screen space.
We then create another variant that directly projects the aggregated LiDAR point cloud without assigning each point a fixed radius in the image screen space.
The results in Table~\ref{tab:ablation_condition} indicate that our model performs better since the projected point cloud has difficulty handling occlusion relationships .

\PAR{Analysis of the distillation process.}
We carry out ablation studies on two sequences from the Waymo~\cite{Sun_2020_CVPR} dataset to analyze our \methodnameblank distillation framework in Table~\ref{tab:ablation_gaussian} and Figure~\ref{fig:ablation_gaussian}.
(1) We set $ \lambda_{\text{lpips}}$ to 0 and change $ \mathcal{L}_{\text{novel}}$ to L1 loss. 
The result in Figure~\ref{fig:ablation_gaussian} indicates that LPIPS loss helps recover sharp details under a novel viewpoint.
(2) We remove the novel weight by setting $ \lambda_{\text{novel}} $ as 1.0. 
There is a significant drop in all the metrics and many artifacts appear on the moving object as shown in Figure~\ref{fig:ablation_gaussian}, which highlights the importance of treating input views and novel views independently.
(3) We set the noise scale $s$ to 1 throughout the training so that the model always starts from a pure noise. Artifacts appear in areas where LiDAR conditions are lacking, such as the road signs on traffic lights.

\subsection{Scene Editing}
\begin{figure}
    \centering
    \includegraphics[width=1.0\linewidth]{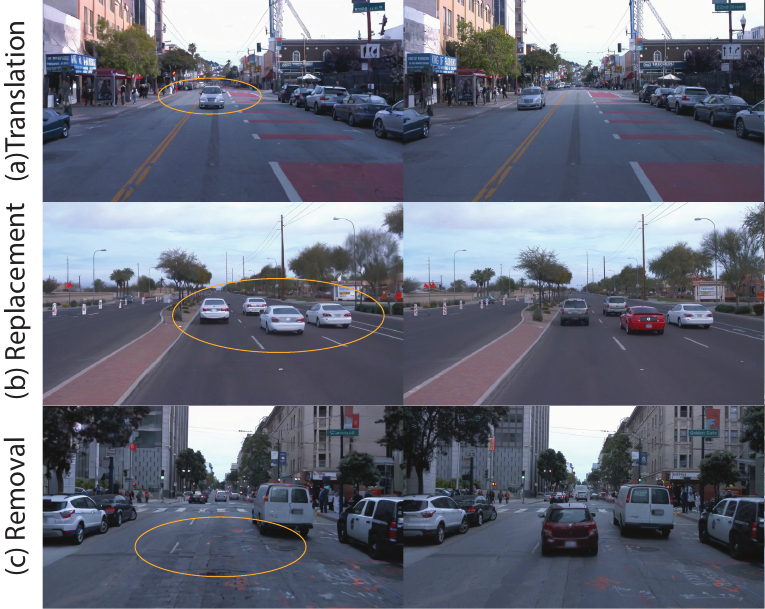}
    \caption{
        \textbf{Editing results on the Waymo \cite{Sun_2020_CVPR} dataset.}
        \methodnameblank supports various editing operations, including translation, replacement, and removal.
        Images in the right and left columns represent the results before and after editing, respectively.
    }
    \label{fig:waymo_editing}
\end{figure}

\methodnameblank supports various editing operations for moving objects.
We can achieve object translation (Figure \ref{fig:waymo_editing} (a)), replacement (Figure \ref{fig:waymo_editing} (b)) and removal (Figure \ref{fig:waymo_editing} (c)) by adjusting the attributes of the object bounding boxes during multi-frame point cloud aggregation 
to provide modified LiDAR conditions for the video diffusion model. In contrast to previous reconstruction methods~\cite{tonderski2024neurad,yang2023unisim,yan2024street}, which model each object separately, \methodnameblank 
can perform editing operations without per-scene optimization.

\section{Conclusion}
This paper introduced \methodname, a controllable video diffusion model for street view synthesis.
The key insight is to leverage sparse yet geometrically accurate LiDAR to provide pixel-level conditions for precise camera control, enabling the model to generate consistent video frames aligned with the camera inputs.
By further distilling \methodname into a 3DGS~\cite{kerbl3Dgaussians} model, we enable real-time view synthesis in challenging scenarios, such as lane shift. 
Moreover, scene editing is possible by providing modified LiDAR conditions to the video diffusion model. 
Detailed ablation and comparisons are conducted on several datasets, demonstrating the effectiveness of the proposed methods. 

This work also has some known limitations. 
First, collecting the necessary LiDAR data and object tracklets for training \methodnameblank is costly due to the extensive data collection and processing requirements.
Second, the inference speed of \methodnameblank remains significantly short of real-time due to 
the architecture of video denoising U-Net. Future work could consider employing a more advanced model for real-time inference.

\PAR{Acknowledgement.} The authors would like to acknowledge the support from NSFC (No. 62402427, U24B20154), Li Auto, Information Technology Center and State Key Lab of CAD\&CG, and Zhejiang University Education Foundation Qizhen Scholar Foundation.

{
    \small
    \bibliographystyle{ieeenat_fullname}
    \bibliography{main}
}

\newpage

\section*{Appendix}
\section{More Implementation Details}
\subsection{\methodnameblank Training Details}
We construct the training video clips using the front camera and LiDAR sensor of Waymo Open~\cite{Sun_2020_CVPR} and PandaSet~\cite{xiao2021pandaset} datasets, with the start frame of each video clip selected at the interval of 0.5 second (5 frames for both dataset).
We set the radius of each LiDAR point cloud in NDC space to 0.01 and crop the upper part of LiDAR condition maps to match the input resolution of the diffusion model during both training and inference.

For adaptation from the pretrained model of Vista~\cite{gao2024vista}, we ignore the action control layer injected via cross-attention and mark the first element of the frame-wise mask to 1 and the rest to 0.
We incorporate the LoRA~\cite{hu2022lora} adapters introduced during the learning of action controllability as it contributes to the enhancement of visual quality~\cite{gao2024vista}.
More details can be found in the original paper. 

During the low-resolution training stage, we sample exclusively from Waymo dataset. 
During the high-resolution training stage, we sample from a hybrid dataset, combining Waymo Open and PandaSet datasets with sampling probabilities of 0.9 and 0.1, respectively.

\subsection{\methodnameblank Distillation Details}
\PAR{Loss function.}
We jointly optimize the gaussian parameters of background and foreground moving objects, texel of the high-resolution sky cubemap and noisy object tracklets following Street Gaussians~\cite{yan2024street}.
The extra loss $\mathcal{L}_{\text{g}}$ for input view camera is defined as:

\begin{equation}
    \label{eq:gaussian_loss_input}
    \mathcal{L}_{\text{g}} = \lambda_\text{depth} \mathcal{L}_{\text{depth}} + \lambda_\text{sky} \mathcal{L}_{\text{sky}} + \lambda_\text{reg} \mathcal{L}_{\text{reg}},
\end{equation}

where $\mathcal{L}_{\text{depth}}$, $\mathcal{L}_{\text{sky}}$ and $\mathcal{L}_{\text{reg}}$ share the same format as Street Gaussians. Please refer to the original paper for more details.
The coefficients $\lambda_\text{depth}$, $\lambda_\text{sky}$ and $\lambda_\text{reg}$ in Equation~\ref{eq:gaussian_loss_input} are set to 0.01, 0.05 and 0.1, respectively.
For the loss function of novel view camera, we crop the upper part of the rendering image and resize to $576 \times 1024$ to compute the LPIPS~\cite{Zhang_2018_CVPR} loss with novel view image generated by \methodnameblank.  



\PAR{Point cloud initialization.}
We initialize the background gaussian model as the combination of LiDAR and SfM point cloud. 
The object gaussian model is initialized with aggregated LiDAR points obtained from object tracklets or random sampling.  
The colors of LiDAR points are assigned by projecting them onto the nearest image plane.

\PAR{Optimization.}
We adopt the densification strategy introduced in~\cite{Yu2024GOF} to prevent suboptimal solutions by accumulating the norms of view-space position gradients.
The densification threshold is set to 0.0006. We disable the pruning of big gaussians in world space since this hinders the gaussian model to represent distant regions and the LiDAR points have provided a good initialization to prevent the model from falling into local optima.  
We finally introduce the 2D Mip filter to enable anti-aliased rendering inspired by~\cite{Yu2023MipSplatting}. 

We sample \methodnameblank every 5000 iterations from iteration 7000 to 22000 and linearly reduce the noise scale $s$ from 0.7 to 0.3. 
We use the annotated object tracklets provided by the datasets, with the learning rates for the translation vector and rotation matrix initialized at $5e^{-4}$ and $1e^{-5}$, respectively, decaying exponentially to $1e^{-5}$ and $5e^{-6}$.
For the remaining parameters, we use the default values from the official implementation of Street Gaussians.

\subsection{Evaluation Details}
\PAR{Interpolation.}
For the interpolation setting of Ours-V in the main paper, we incorporate training images along the input trajectory in addition to the reference image and LiDAR conditions.
During each denoising step, we replace the prediction of $\mathcal{F}_\theta$ at training frame with the clean latent of training images. 
This could lead to improvement in the interpolation quality, with PSNR increasing from 23.66 to 27.19 and LPIPS decreasing from 0.098 to 0.087 on Waymo Open Dataset~\cite{Sun_2020_CVPR}.

\PAR{Baselines.}
We use the same object tracklets as our method for all the baselines requiring 3D bounding box as input~\cite{yan2024street,tonderski2024neurad,yang2023unisim}.
We use the same rendering kernel and optimization strategy as our method for all the baselines using 3DGS as the scene representation~\cite{kerbl3Dgaussians,yan2024street}.

\PAR{Metric.} For Fréchet inception distance (FID) metric, we mark the input video as real and the rendered sequence as unreal for each scene.    
For Ours-V, we upper-crop the input video to match the resolution of the generated video.

\section{Additional Experiments}

\subsection{More Comparisons}
\begin{figure*}
    \centering
    \includegraphics[width=1\linewidth]{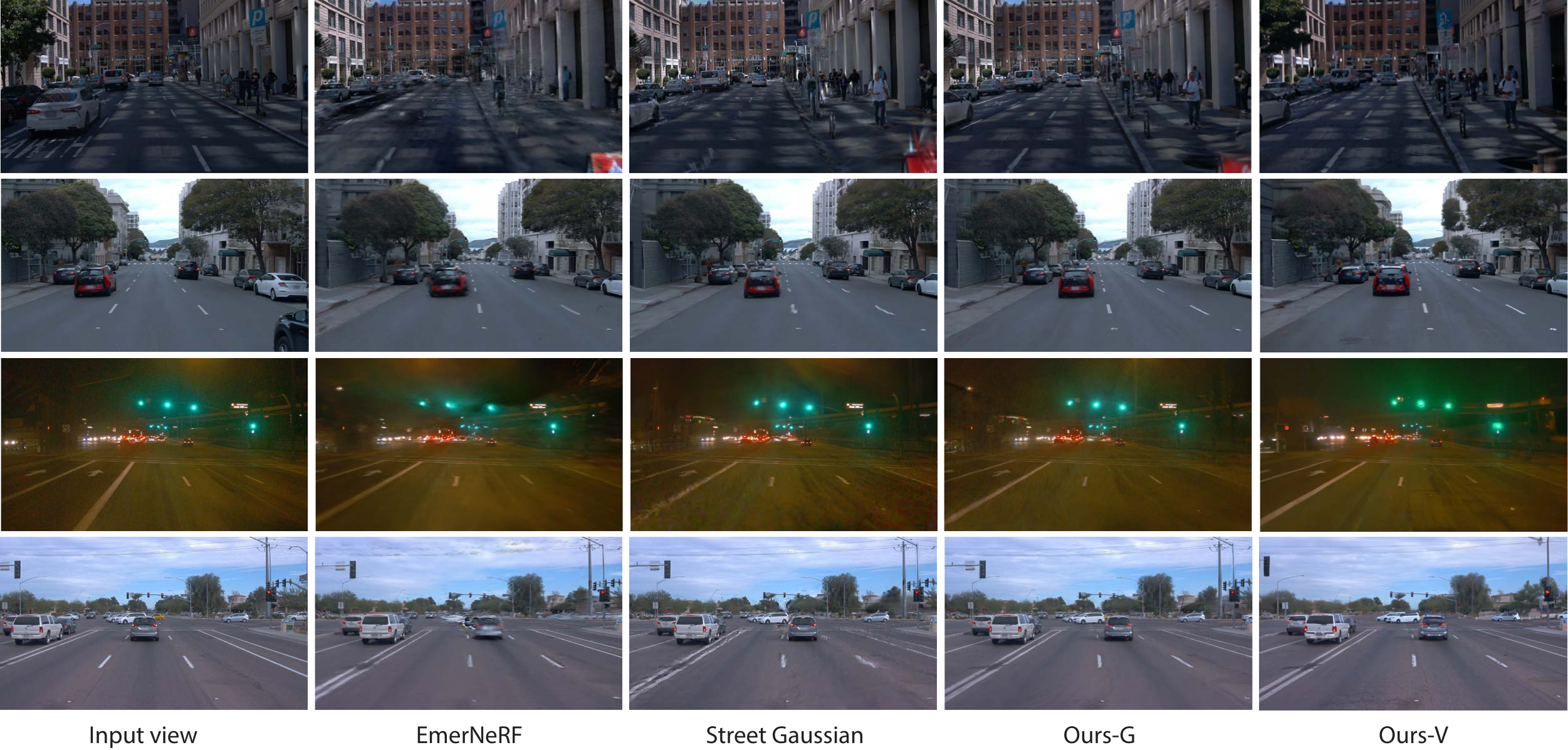}
    \caption{
        \textbf{Qualitative comparisons on the Waymo~\cite{Sun_2020_CVPR} dataset. } The camera is gradually laterally shifted for 3 meters to left or right.
    }
    \label{fig:waymo_comparision_supp}
\end{figure*}

\begin{figure*}
    \centering
    \includegraphics[width=1\linewidth]{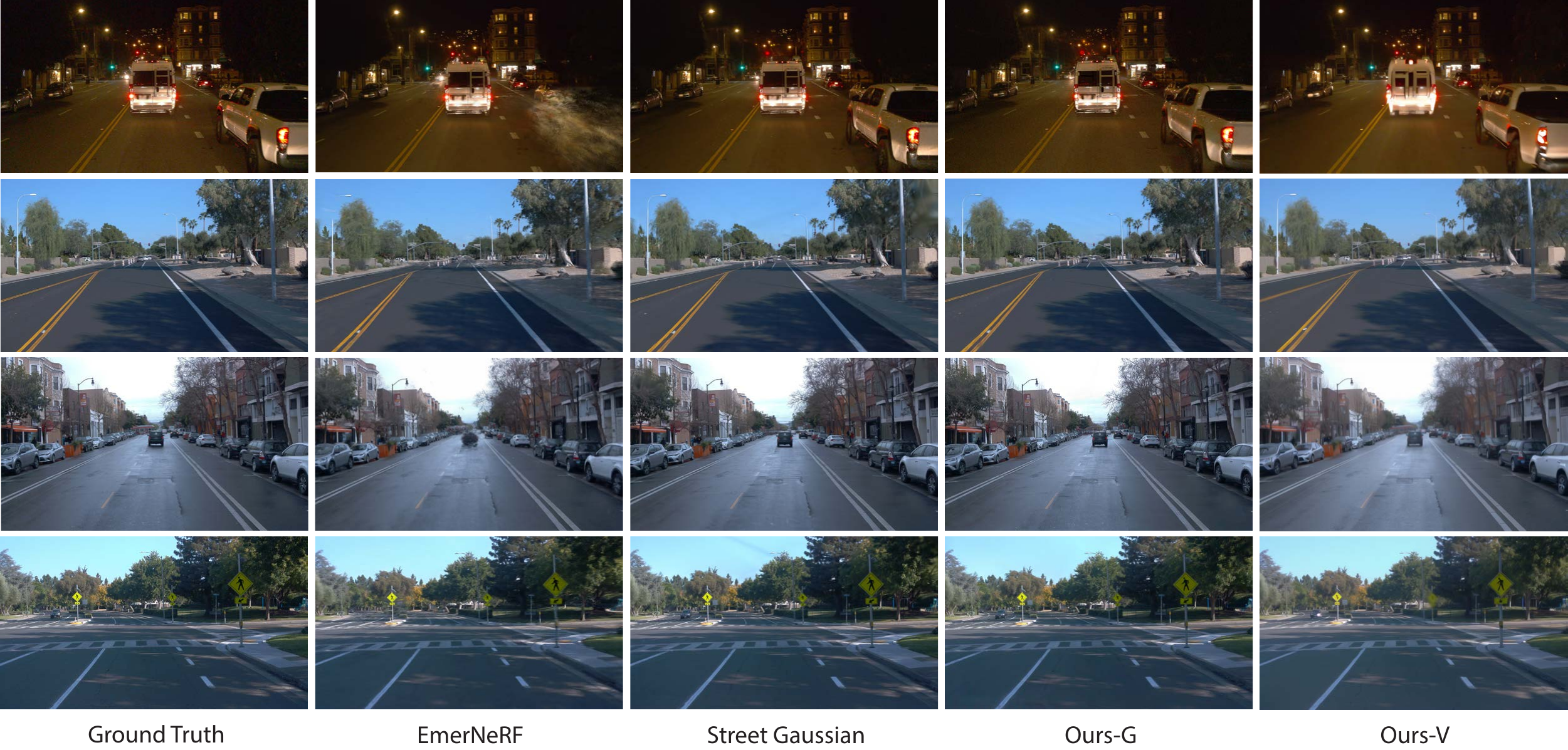}
    \caption{
        \textbf{Qualitative comparisons of view interpolation on the Waymo~\cite{Sun_2020_CVPR} dataset}.
    }
    \label{fig:waymo_interpolation}
\end{figure*}

\begin{figure*}
    \centering
    \includegraphics[width=1\linewidth]{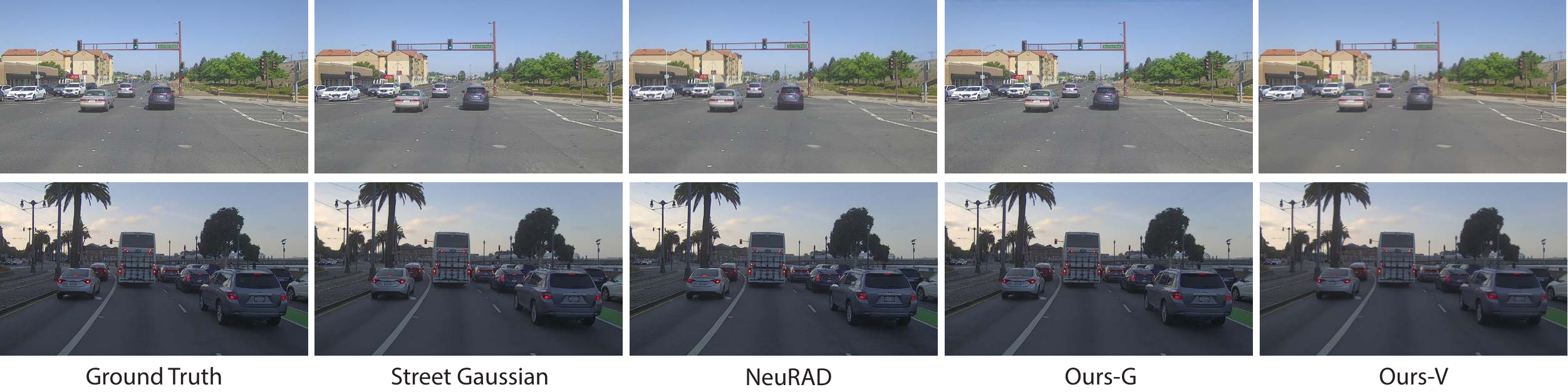}
    \caption{
        \textbf{Qualitative comparisons of view interpolation on the PandaSet~\cite{xiao2021pandaset} dataset}.
    }
    \label{fig:pandaset_interpolation}
\end{figure*}

\PAR{Comparisons with baselines.}
We provide more qualitative comparisons on Waymo~\cite{Sun_2020_CVPR} dataset under the setting of lane change in Figure~\ref{fig:waymo_comparision_supp}. 
Figures~\ref{fig:waymo_interpolation}, \ref{fig:pandaset_interpolation} display the view interpolation results on Waymo Open~\cite{Sun_2020_CVPR} and PandaSet~\cite{xiao2021pandaset} datasets.
Our method achieves comparable rendering quality to the baselines under input trajectory while achieving significantly better results for view extrapolation.

\PAR{Comparisons with concurrent works.}
We compare Ours-V with ViewCrafter~\cite{yu2024viewcrafter} and Ours-G with DriveDreamer4D~\cite{zhao2024drive}, both of which are concurrent of our works.
For ViewCrafter, we make several modifications to improve its performance. 
First, we build the global point cloud from the whole input sequence instead of selecting the first frame as in the original setting.
Second, we fix the camera parameters during the global alignment process of DUSt3R~\cite{wang2024dust3r} by using camera calibration results from the dataset. 
This can also help define camera poses within the dataset's coordinate system when performing view extrapolation.
For DriveDreamer4D, we compare our method on segment-103593 of Waymo following their setting with PVG \cite{chen2023periodic} as the base model.   

\begin{figure}[!htbp]
    \centering
    \includegraphics[width=\linewidth]{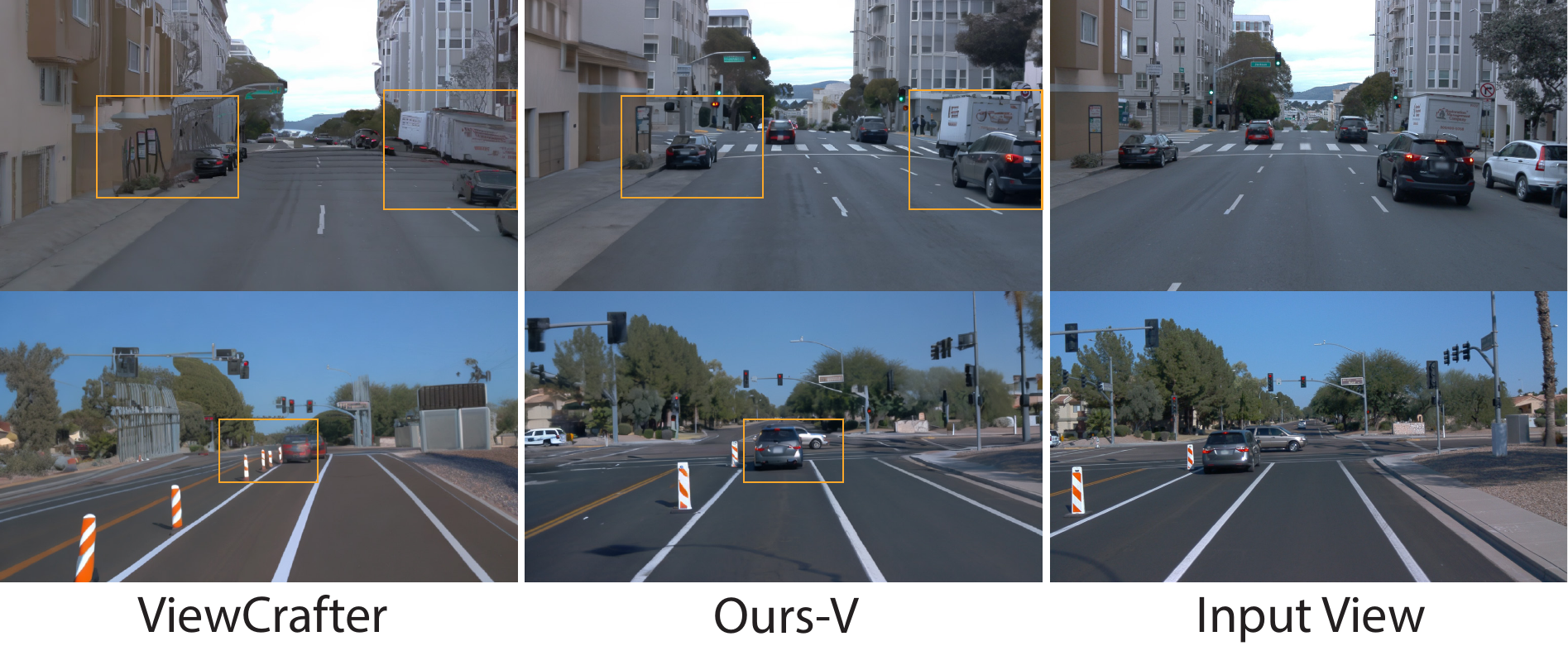}
    \caption{Qualitative comparison with ViewCrafter \cite{yu2024viewcrafter} under novel trajectory. The camera is laterally shifted for 3 meters. Input view denotes the closest input video frames.}
    \label{fig:viewcrafter_novel}
  \end{figure}

\begin{figure}[!htbp]
    \centering
    \includegraphics[width=\linewidth]{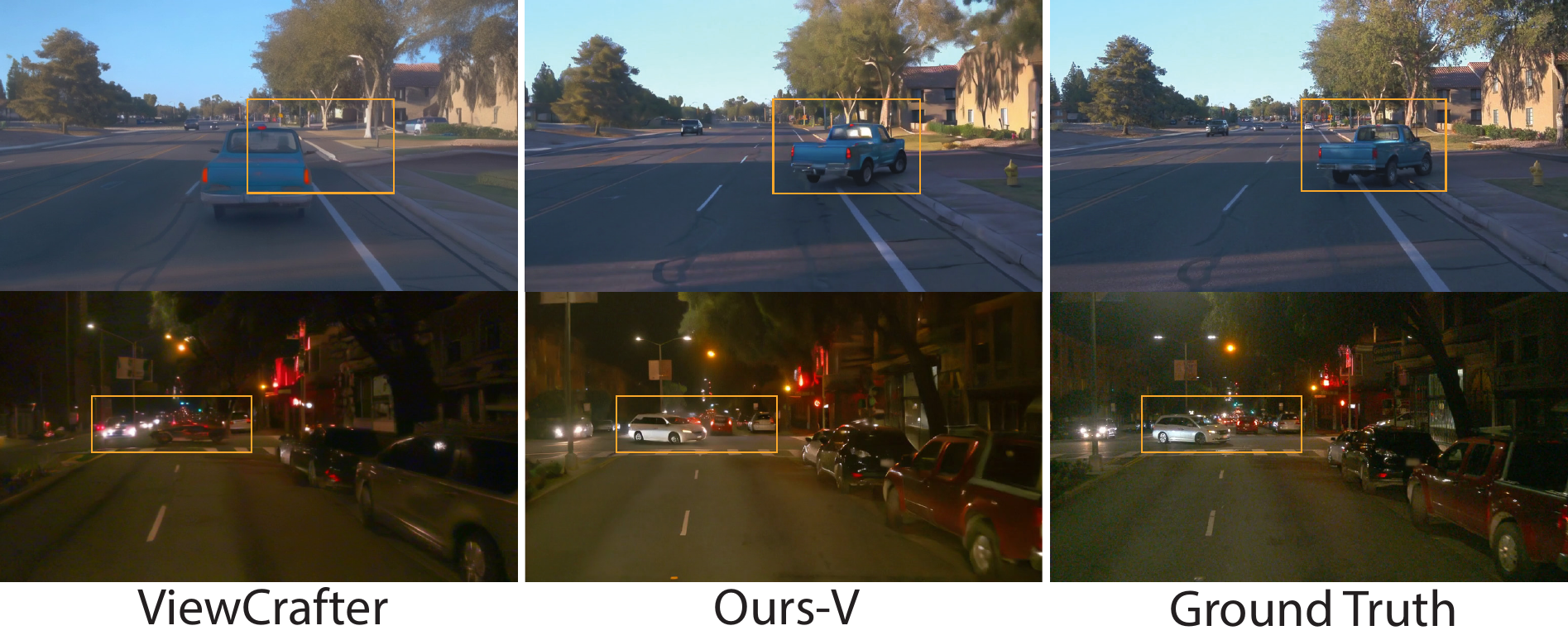}
    \caption{Qualitative comparison with ViewCrafter \cite{yu2024viewcrafter} under input trajectory. Our model can handle moving objects.}
    \label{fig:viewcrafter_input}
  \end{figure}

As shown in Table \ref{tab:viewcrafter} and Figures \ref{fig:viewcrafter_novel}, \ref{fig:viewcrafter_input}, 
our method achieves better view synthesis results under both input and novel trajectory compared with ViewCrafter \cite{yu2024viewcrafter}. 
ViewCrafter builds point cloud without considering that the geometry of dynamic scene changes overtime, thus it fails to accurately model dynamic regions as shown in Figure \ref{fig:viewcrafter_input}.
The generated results also degrade significantly as the camera deviates from the input trajectory since the predicted point cloud is defined in the camera coordinate system.
In contrast, the LiDAR point cloud defined in the world coordinate system provides our model with stronger generalization ability on new trajectories even if no ground truth data is available during training.

\begin{figure}[!htbp]
    \centering
    \includegraphics[width=\linewidth]{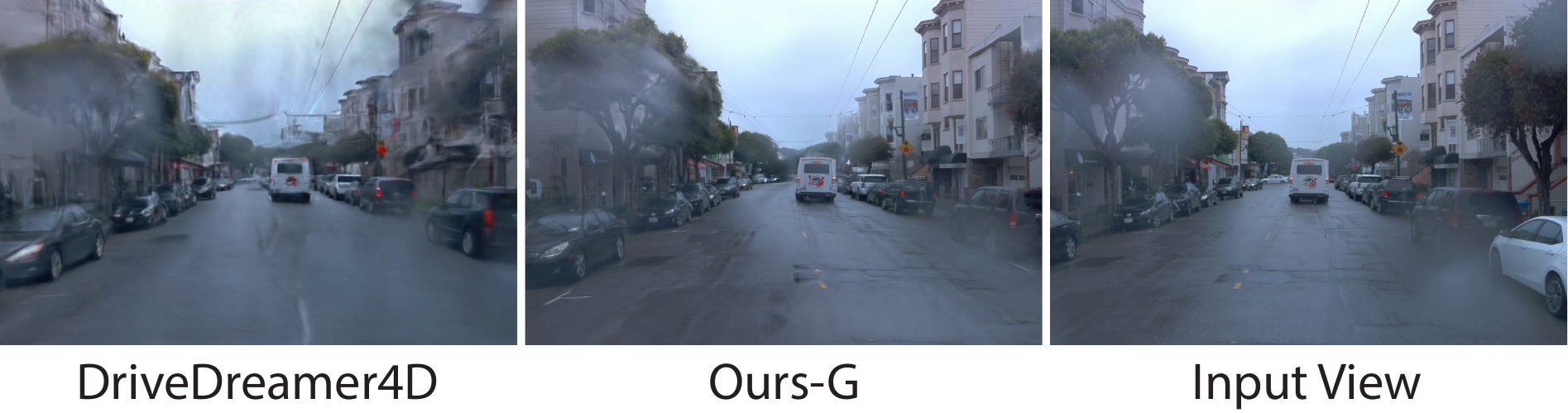}
    \caption{Qualitative comparison with DriveDreamer4D \cite{zhao2024drive} under novel trajectory. The camera is laterally shifted for 2 meters.}
    \label{fig:drivedreamer4d}
  \end{figure}

As shown in Figure~\ref{fig:drivedreamer4d}, our method achieves better results under novel trajectory compared with DriveDreamer4D \cite{zhao2024drive} (FID @ 2m 89.71 vs. 91.23, FID @ 3m 96.11 vs. 123.32).
Due to the sparse conditioning of DriveDreamer4D, the generated videos often lack accurate 3D perception. 
This leads to noticeable artifacts in the reconstructed scene, such as the vehicle in the lower left corner and the building on the right side.

\begin{table}[ht]
    \centering
    \scalebox{0.7}{
    \begin{tabular}{ccccccc}
        \toprule
        \multirow{2}{*}{Methods} & \multicolumn{3}{c}{Input trajectory} & \multicolumn{2}{c}{Novel trajectory} \\
        \cmidrule(lr){2-4} \cmidrule(lr){5-6}
        & PSNR$\uparrow$ & LPIPS$\downarrow$ & FID $\downarrow$ & FID$\downarrow$ @ 2m & FID$\downarrow$ @ 3m \\
        \midrule
        ViewCrafter \cite{yu2024viewcrafter} &  21.59 &  0.226 & 97.86 & 135.69 & 137.76     \\
        Ours-V & \textbf{25.90} & \textbf{0.143} & \textbf{60.49}& \textbf{62.43} & \textbf{73.49} \\
        \bottomrule
    \end{tabular}
    }
    \caption{
        Quantitative comparison with ViewCrafter \cite{yu2024viewcrafter}. We use the video clips in ablation to test the results on input trajectory and the scenes in experiment to test the results on novel trajectory.
        Metrics are averaged over all sampled sequences.
    }
    \label{tab:viewcrafter}
\end{table}

\subsection{Adapting to Advanced Video Diffusion Model}
We can further enhance the performance of \methodname by adapting it to more advanced video diffusion models, such as Wan~\cite{wan2025}.
To be specific, we adopt Wan2.1-14B-I2V as our base model and patchify the video and LiDAR condition latents separately with the same embedding layer. 
The LiDAR condition latents are first passed through a zero-initialized linear layer and then element-wise added to the video latents before being fed into the diffusion transformer blocks.
During training, instead of optimizing all the parameters as in Vista, we utilize a Low-Rank Adaptation approach (LoRA)~\cite{hu2022lora} for the diffusion transformer
while keeping the pretrained parameters frozen to improve the training efficiency and preserve the 2D video prior.

\begin{figure*}
    \centering
    \includegraphics[width=1\linewidth]{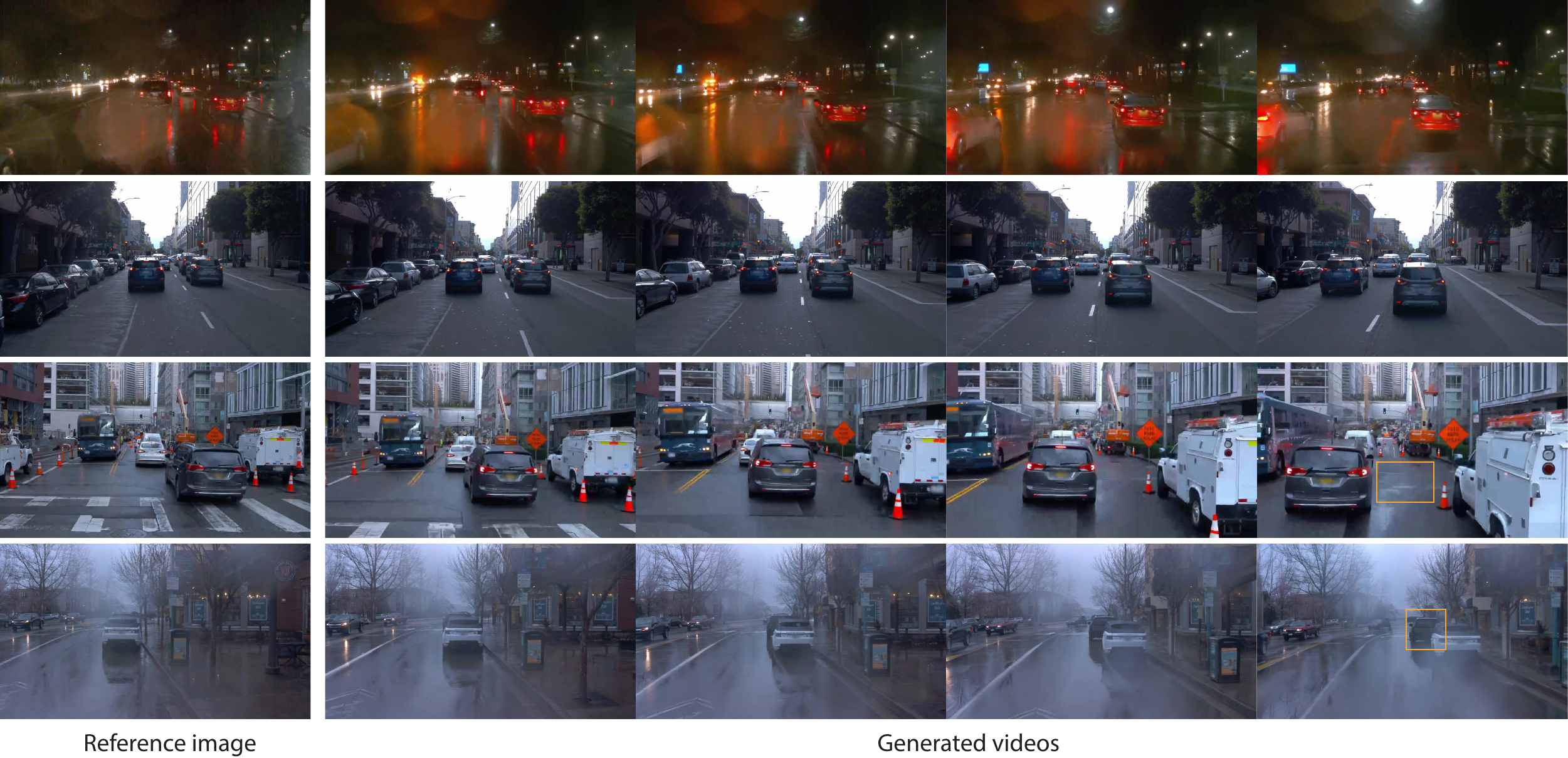}
    \caption{
        \textbf{Visual results of \methodname based on Wan2.1~\cite{wan2025}.} When having a more powerful video generation model, 
        \methodname is able to produce realistic results under challenging scenes with severe occlusions and diverse weather conditions. 
        The camera is gradually shifted for 3 meters to left or right.
    }
    \label{fig:wan_video}
\end{figure*}

As shown in Figure~\ref{fig:wan_video}, \methodname based on Wan can generate high quality results even under extreme viewpoint changes. 
For instance, the regions highlighted in Figure~\ref{fig:wan_video} are completely invisible to the input camera and LiDAR sensors, which means no condition is available for these areas 
during inference. Nevertheless, our method can still produce plausible results thanks to the powerful generative prior of the base model. 
This demonstrates a key strength of our method: it can be readily adapted to more advanced video diffusion models, achieving improved performance with minimal effort.

\subsection{More Editings}
\begin{figure*}
    \centering
    \includegraphics[width=1.0\linewidth]{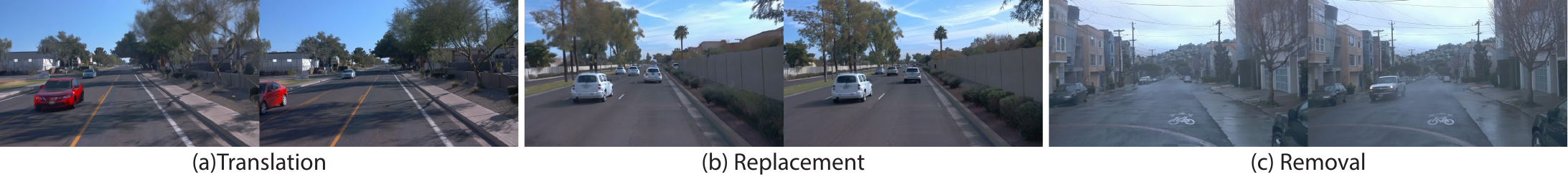}
    \caption{
        \textbf{More editing results on the Waymo \cite{Sun_2020_CVPR} dataset.}
        The right and left images represent the results before and after editing.
    }
    \label{fig:waymo_editing_supp}
\end{figure*}

We provide more visual results of scene editing in Figure~\ref{fig:waymo_editing_supp} including object translation, replacement and removal.

\subsection{More Ablations}

\begin{figure*}
    \centering
    \includegraphics[width=1\linewidth]{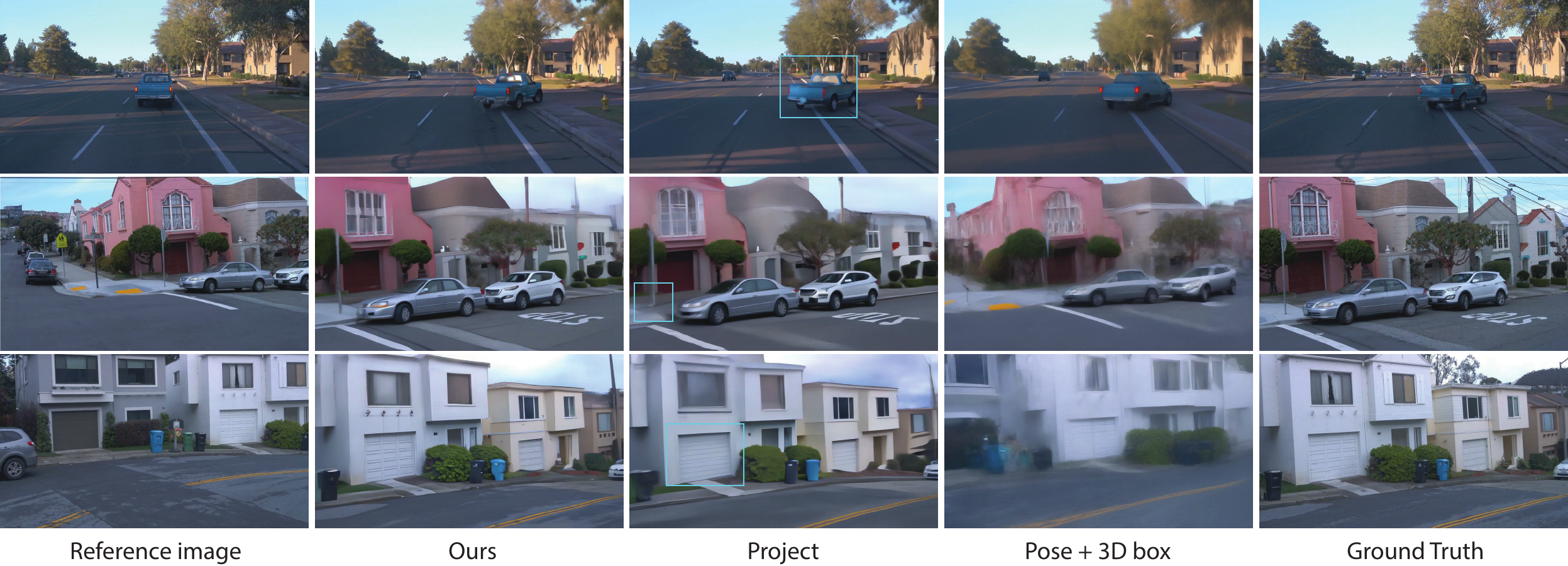}
    \caption{
        \textbf{Visual ablation results on the design choice of \methodname.} 
    }
    \label{fig:ablation_streetcrafter_supp}
\end{figure*}

\paragraph{Analysis of LiDAR conditions}
We present more visual comparisons of the design choice of \methodnameblank in Figure~\ref{fig:ablation_streetcrafter_supp}.
The generated frames under the guidance of camera parameter as vector are blurry when the target viewpoint move away from the reference image.  
Although the 3D bounding box can provide priors regrading object motions, it still fails to align well with the target image as shown in the first row of Figure~\ref{fig:ablation_streetcrafter_supp}.
The results under the condition of projected multi-frame LiDAR can preserve the scene structure but still lack details in regions with rich texture.
\paragraph{Analysis of the novel view sampling ratio}
We conduct experiment on one Waymo sequence to analyze the influence of novel view sampling ratio $p$.
The results in Table~\ref{tab:ablation_sample_ratio} indicates that $p=0.4$ yields the overall best result.
\begin{table}[ht]
    \centering
    \scalebox{0.8}{
    \begin{tabular}{lccccc}
        \toprule
        \multirow{2}{*}{Methods} & \multicolumn{2}{c}{Interpolation} & \multicolumn{2}{c}{Lane Shift} \\
        \cmidrule(lr){2-3} \cmidrule(lr){4-5}
        & PSNR$\uparrow$ & LPIPS$\downarrow$ & FID$\downarrow$ @ 2m & FID$\downarrow$ @ 3m \\
        \midrule
        (1) $p = 0.8$ & 28.76 & 0.059 & 72.76 & 84.78 \\ 
        (2) $p = 0.6$ & 29.61 & 0.049 & 68.33 & 81.26 \\
        (3) $p = 0.4$ & \textbf{30.42} & \textbf{0.041} & 67.54 & \textbf{79.19} \\
        (4) $p = 0.2$ & 30.29 & \textbf{0.041} & \textbf{67.09} & 81.26 \\
        \bottomrule
    \end{tabular}
    }
    \caption{
        Ablations on the novel view sampling ratio $p$.  
    }
    \label{tab:ablation_sample_ratio}
\end{table}


\paragraph{Analysis of the noise scale}
We conduct experiment on one Waymo sequence to analyze the influence of noise scale $s$.
We have demonstrated in the main paper that adding noise to the render latents leads to better scene consistency than starting from gaussian noise. 
Since the added noise would have little influence when $s$ is less than 0.3 according to the sampling scheme of Vista~\cite{Karras2022edm}, we set $s_\text{min}$ to 0.3 and ablate on the value of $s_\text{max}$. 
The results in Table~\ref{tab:ablation_noise_scale} indicates that reducing $s$ from 0.7 to 0.3 maintains a balance between sampling steps and rendering quality.

\begin{table}[ht]
    \centering
    \scalebox{0.7}{
    \begin{tabular}{lccccc}
        \toprule
        \multirow{2}{*}{Methods} & \multicolumn{2}{c}{Interpolation} & \multicolumn{2}{c}{Lane Shift} \\
        \cmidrule(lr){2-3} \cmidrule(lr){4-5}
        & PSNR$\uparrow$ & LPIPS$\downarrow$ & FID$\downarrow$ @ 2m & FID$\downarrow$ @ 3m \\
        \midrule
        (1) $s_\text{max}=1.0, s_\text{min}=0.3$ & 30.08 & 0.044 & 69.66 & 79.87 \\ 
        (2) $s_\text{max}=0.7, s_\text{min}=0.3$  & 30.42 & \textbf{0.041} & \textbf{67.54} & \textbf{79.19} \\
        (3) $s_\text{max}=0.5, s_\text{min}=0.3$ & \textbf{30.46} & 0.042 & 68.68 & 81.23 \\
        \bottomrule
    \end{tabular}
    }
    \caption{
        Ablations on the noise scale $s$.  
    }
    \label{tab:ablation_noise_scale}
\end{table}

\subsection{Deformable Objects}
\begin{figure*}
    \centering
    \includegraphics[width=1\linewidth]{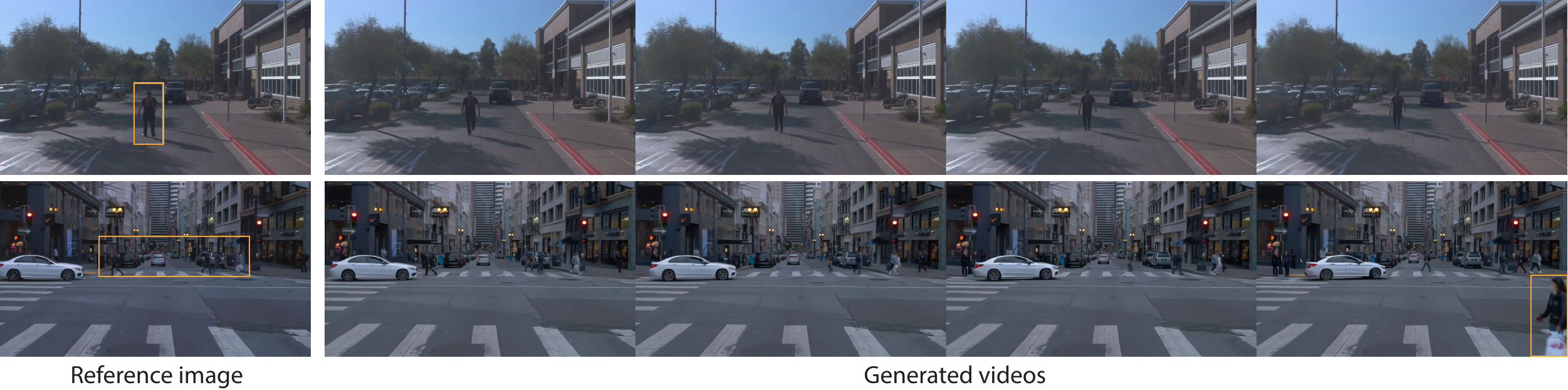}
    \caption{
        \textbf{Visual results of \methodname for scene with deformable objects}.
    }
    \label{fig:deformable}
\end{figure*}

We show the generated videos of \methodnameblank under scenes with multiple deformable objects such as pedestrians in Figure~\ref{fig:deformable}.
Although multi-frame LiDAR aggregation leads to incorrect guidance for deformable objects, our method can produce plausible results thanks to  
the generative prior of video diffusion model.

\afterpage{\clearpage}
\onecolumn
\twocolumn


\end{document}